\documentclass[lettersize,journal]{IEEEtran}

\usepackage{amsmath,amsfonts,amssymb}
\usepackage{mathtools}
\usepackage{algorithm}
\usepackage{algorithmic}
\usepackage{array}
\usepackage[caption=false,font=normalsize,labelfont=sf,textfont=sf]{subfig}
\usepackage{textcomp}
\usepackage{stfloats}
\usepackage{url}
\usepackage{graphicx}
\usepackage{cite}
\usepackage{booktabs}
\usepackage{multirow}
\usepackage{makecell}
\usepackage{microtype}
\usepackage{bbding}
\usepackage{color}
\usepackage{xcolor}

\graphicspath{{images/}}

\begin{document}

\title{DRDN: Decoupled Representation Dynamic Network for From-Scratch ViT Class-Incremental Learning}

\author{Bingchen~Huang\textsuperscript{*},
        Yifu~Chen\textsuperscript{*},
        Zhiling~Wang\textsuperscript{*},
        and~Yuanchao~Du%
\thanks{\textsuperscript{*}Bingchen Huang, Yifu Chen, and Zhiling Wang contributed equally to this work and share first authorship.}
\thanks{B. Huang, Y. Chen, Z. Wang, and Y. Du are with the Meituan Vision AI Department, Beijing 100190, China (e-mail: huangbingchen@meituan.com; chenyifu05@meituan.com; zhilingwang9681@gmail.com; duyuanchao@meituan.com).}
\thanks{This work has been submitted to the IEEE for possible publication. Copyright may be transferred without notice, after which this version may no longer be accessible.}}

\markboth{Preprint --- Under Review}%
{Huang \MakeLowercase{\textit{et al.}}: DRDN: Decoupled Representation Dynamic Network for Class-Incremental Learning}

\maketitle

\begin{abstract}
Dynamic expansion methods for class-incremental learning (CIL) protect task-specific knowledge by growing dedicated tokens or subnetworks, yet our analyses suggest that classification supervision alone does not sufficiently preserve task-agnostic shared backbone representations over long incremental sequences. We identify two intertwined challenges: \emph{cross-task confusion} from sequential training on predominantly current-task data, which biases decision boundaries toward recent tasks; and \emph{under-optimized shared representations} in the backbone that cap long-term discriminability as tasks accumulate.

We propose the \emph{Decoupled Representation Dynamic Network} (DRDN), which addresses these challenges via two orthogonal mechanisms. For shared backbone representations, DRDN continuously applies masked image modeling (MIM) at every incremental step, with reconstruction gradients routed \emph{exclusively through the backbone}, encouraging it to retain general visual structure beyond class-discriminative cues. For task-specific discrimination, DRDN employs hierarchical task token expansion across all transformer layers, with a modified per-task attention rule that reduces inter-task interference. We support this design with accuracy degradation analysis and cross-task confusion rate measurements.

In the from-scratch ViT CIL setting (no external pretraining), DRDN consistently improves over strong token-expansion baselines with comparable backbone scale. On CIFAR100-B0 (10 steps), DRDN achieves 77.19\% average accuracy, outperforming DKT by 1.36 points and DyTox by 3.53 points, with an advantage that grows at longer incremental sequences. Multi-seed validation confirms stability ($\pm$0.31\%). The MIM decoder is active only during training, adding no inference-time parameters or computation.
\end{abstract}


%%% ============================================================
\section{Introduction}
\label{sec:introduction}
%%% ============================================================

\IEEEPARstart{I}{n} class-incremental learning (CIL), a model must learn new classes sequentially while retaining knowledge of all previously seen classes --- a challenge where standard gradient descent causes \emph{catastrophic forgetting}~\cite{1989,goodfellow-forgetting}. Dynamic expansion methods~\cite{DER,Dytox,DKT} currently achieve leading performance by growing dedicated task-specific components --- tokens or subnetworks --- at each step. While these components protect task-specific knowledge, they leave a critical question unaddressed: \emph{what happens to the shared backbone as tasks accumulate?}

We observe that in dominant ViT-based token-expansion methods (DyTox~\cite{Dytox}, DKT~\cite{DKT}), the shared backbone is trained exclusively through classification objectives that favor task-specific discriminability. This concentration of supervision on task-specific modules creates two intertwined challenges, which we diagnose and validate empirically via accuracy degradation analysis and cross-task confusion analysis (Section~\ref{sec:analysis}).

\begin{figure*}[!t]
  \centering
  \subfloat[Task 0 only (classes 0--9)]{\includegraphics[width=0.32\linewidth]{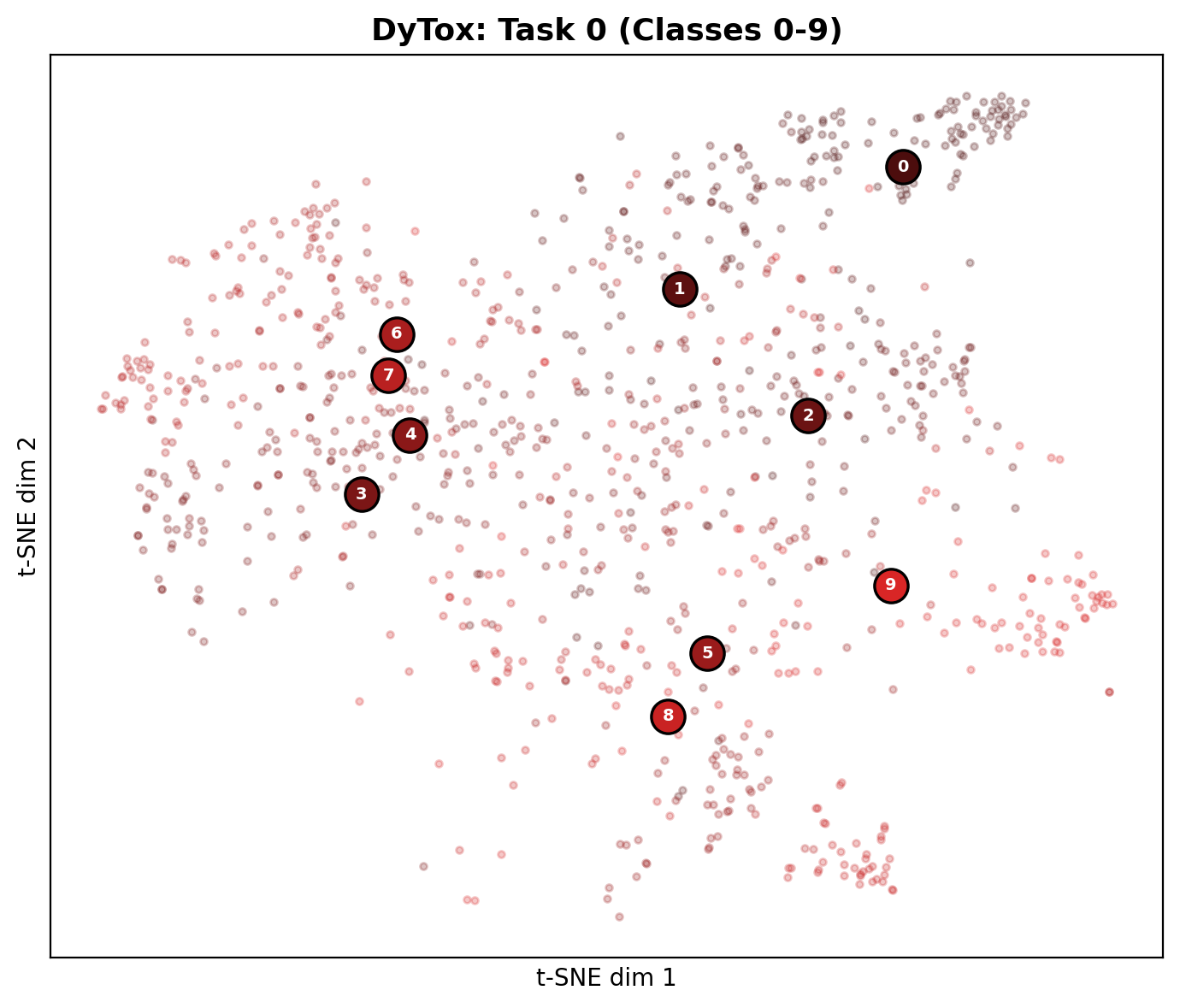}%
  \label{fig:tsne_task0}}
  \hfil
  \subfloat[Task 1 only (classes 10--19)]{\includegraphics[width=0.32\linewidth]{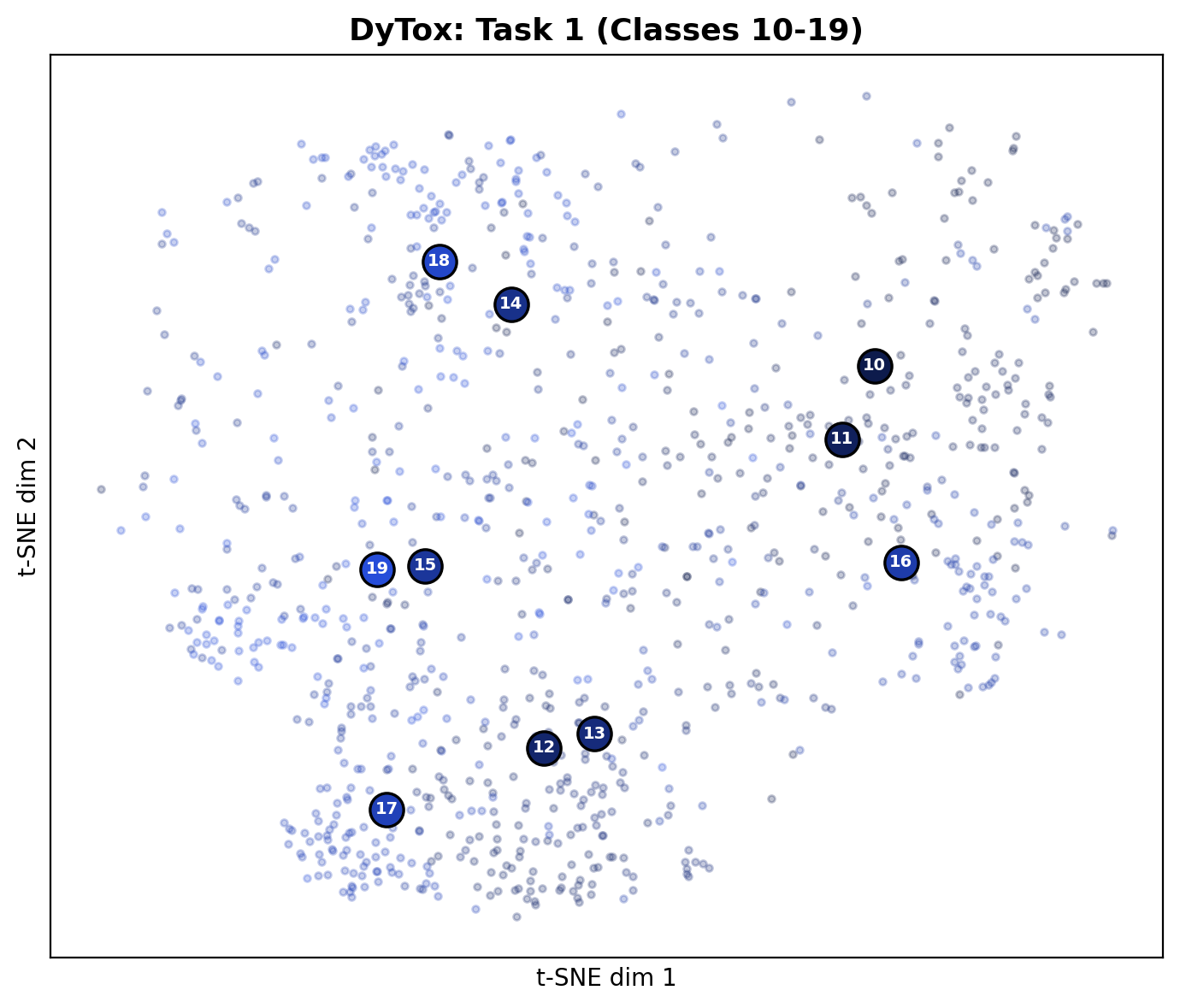}%
  \label{fig:tsne_task1}}
  \hfil
  \subfloat[Tasks 0+1 combined]{\includegraphics[width=0.32\linewidth]{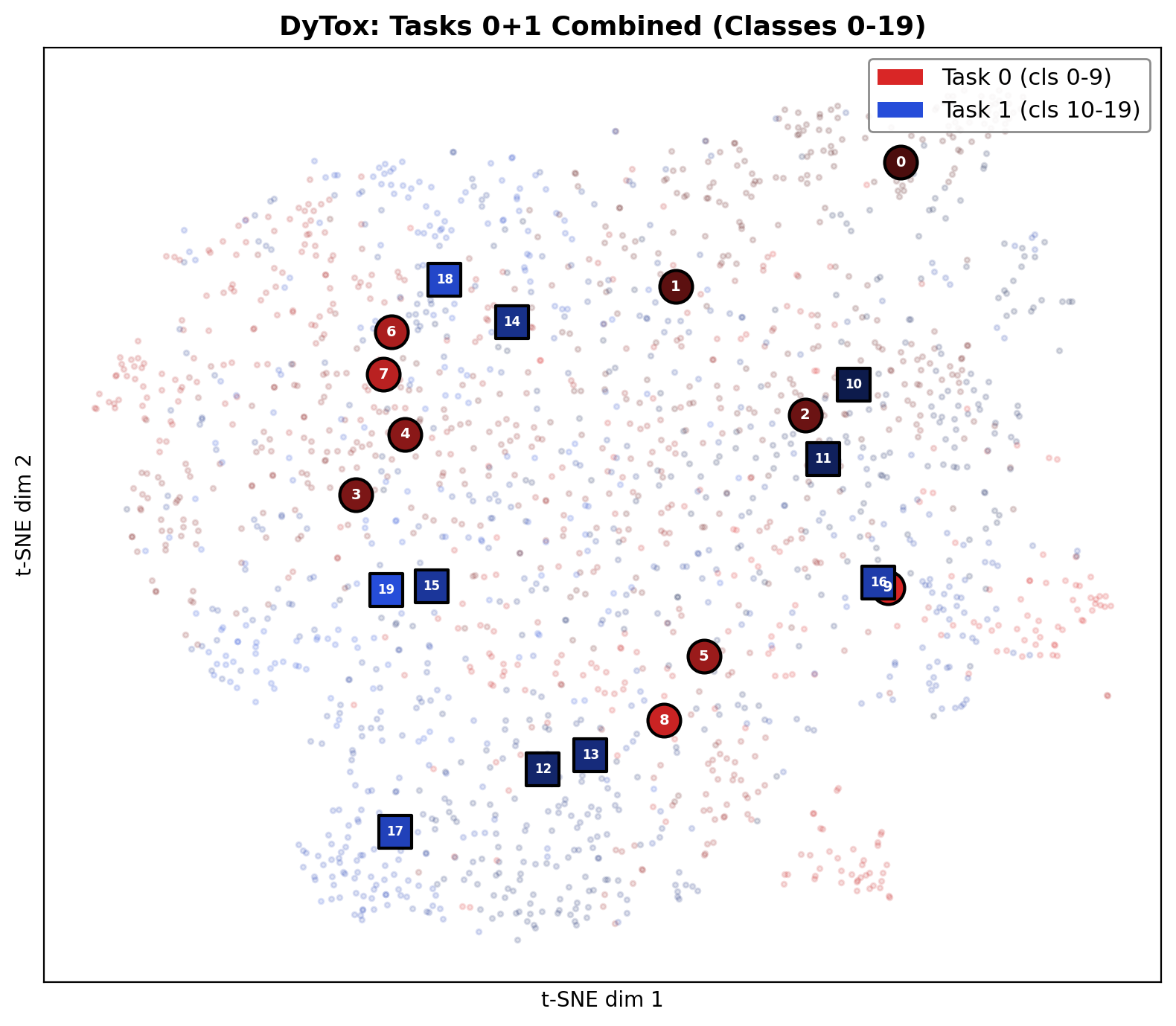}%
  \label{fig:tsne_combined}}
  \caption{t-SNE visualization of DyTox task-token features on CIFAR100-B0 (10 steps) after all 10 tasks are trained. Large markers show per-class centroids; shading within each color family distinguishes individual classes. Within a single task (a, b), class centroids are reasonably spread. When both tasks are overlaid (c), red (Task~0) and blue (Task~1) centroids intermix --- confirming that cross-task confusion is the dominant error mode (90.4\% of errors).}
  \label{fig:task_confusion}
\end{figure*}

\textbf{Challenge 1 (Cross-task confusion).}
Sequential training on predominantly current-task data biases decision boundaries toward recent tasks. During joint inference over all seen classes, semantically similar classes from different tasks are easily confused. As shown in Fig.~\ref{fig:task_confusion}, within a single task (left, right), DyTox class clusters are reasonably separated; when two tasks are plotted together (center), inter-task class boundaries collapse --- different-task classes overlap severely. While prior work has addressed task confusion in ResNet-based expansion methods~\cite{huang2023resolving,BEEF}, we show it persists in ViT token-expansion methods and measure it directly: 90.4\% of DyTox classification errors are cross-task misclassifications (Section~\ref{sec:analysis}).

\begin{figure}[!t]
  \centering
  \includegraphics[width=\linewidth]{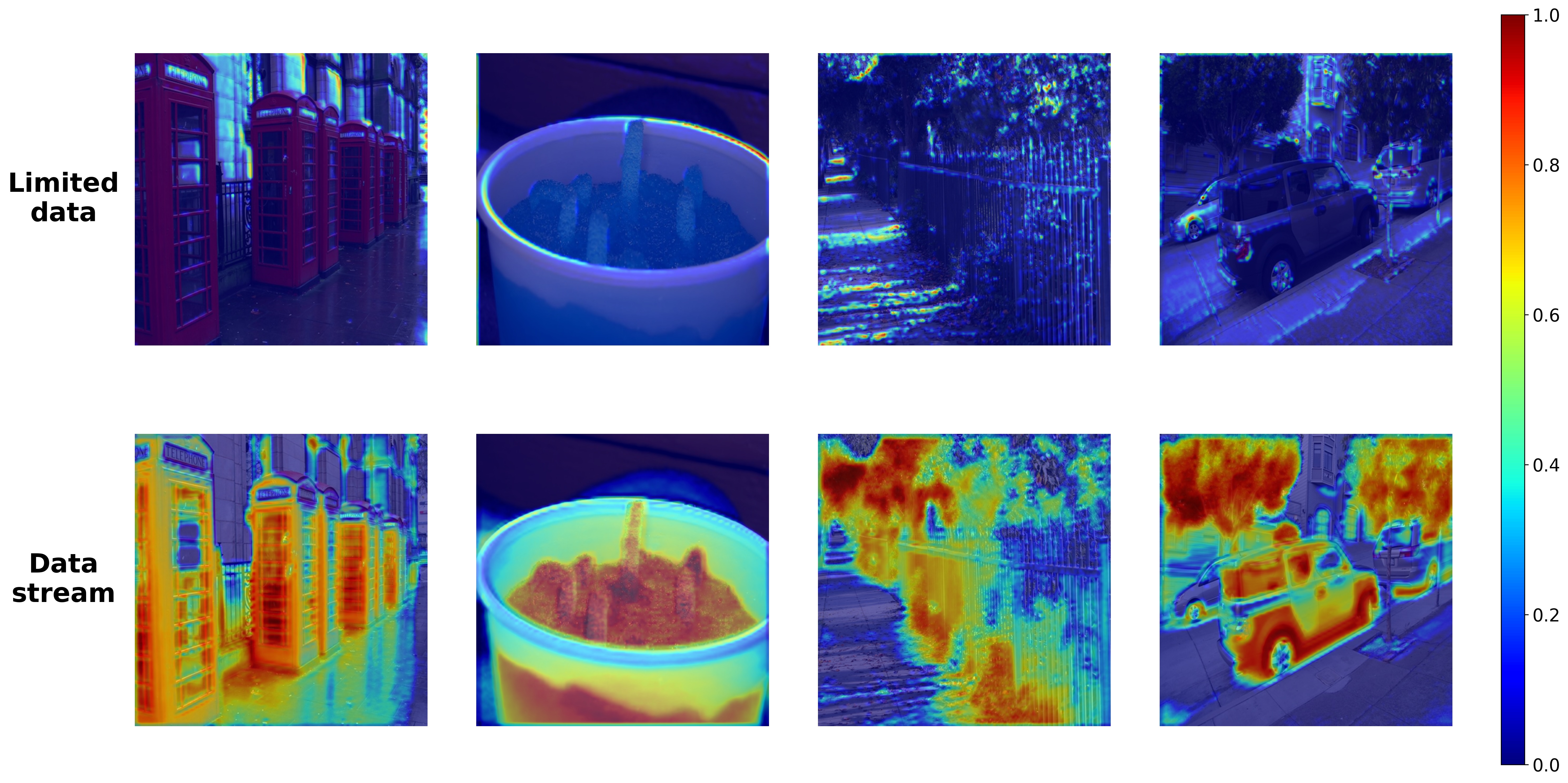}
  \caption{Grad-CAM visualization of shallow-layer activations. Models trained on limited incremental data (left) show diffuse, task-specific activations, while models trained on broader data (right) develop sharp, structurally-grounded activations --- supporting the claim that CIL training under-optimizes shared backbone representations.}
  \label{fig:gradcam}
\end{figure}

\textbf{Challenge 2 (Under-optimized shared representations).}
Shallow ViT layers encode transferable visual structure~\cite{huang2021semantic} that benefits all tasks, yet discriminative training pushes these layers toward task-specific features. As shown in Fig.~\ref{fig:gradcam}, models trained on limited incremental task data develop weaker shallow-layer representations (less precise activation maps on structural cues) compared to models exposed to broader data streams. This manifests as accelerating accuracy degradation: DyTox loses 34.8 absolute points over 10 incremental steps, while DRDN loses only 29.5 points over the same sequence.

These two challenges are potentially linked: weaker shared representations reduce the feature quality that task tokens operate on, which may amplify confusion; and cross-task gradient interference during token learning may in turn pollute shared backbone features. Our analyses suggest both effects are present and that addressing only one leaves substantial room for improvement.

\begin{figure}[!t]
  \centering
  \includegraphics[width=\linewidth]{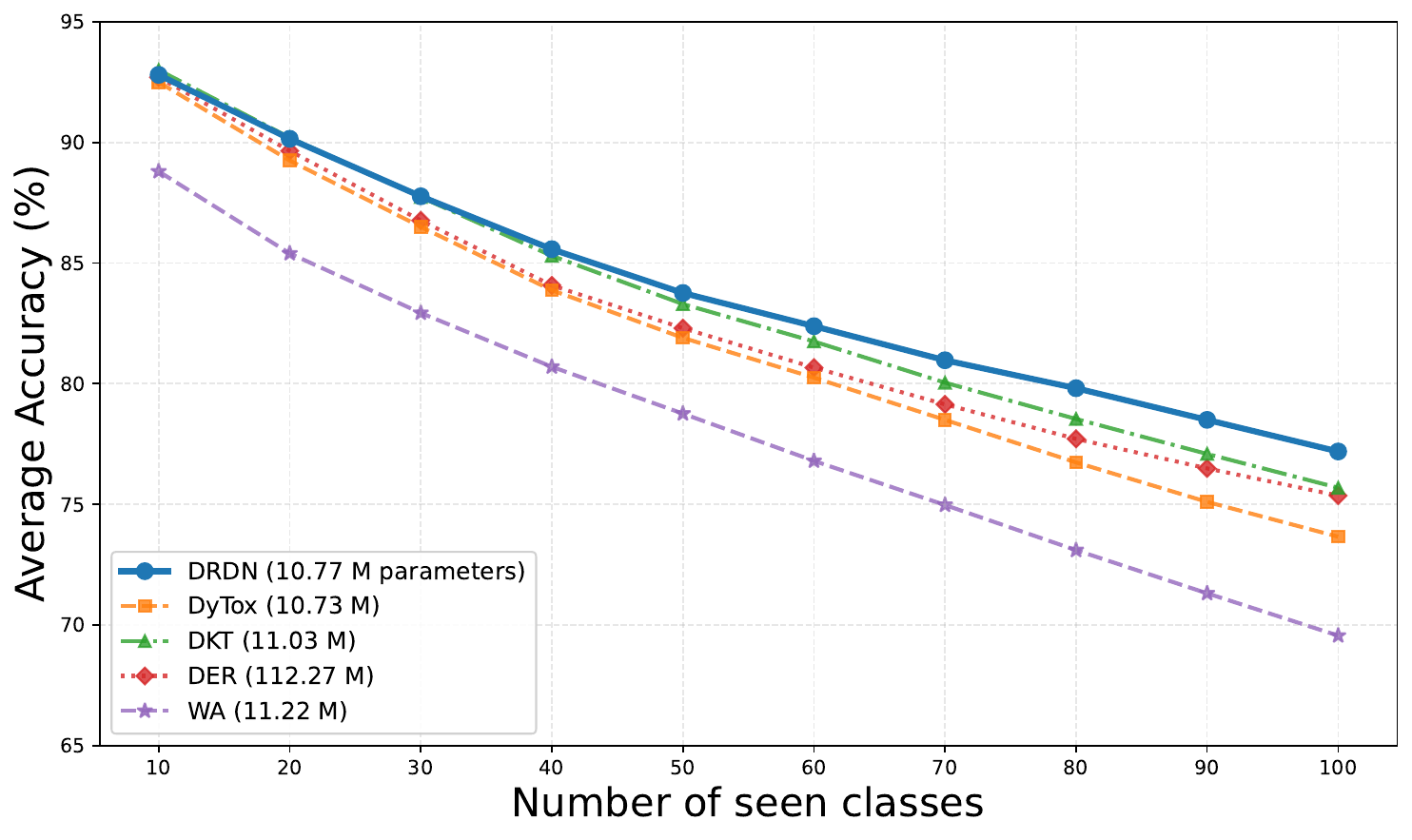}
  \caption{CIL performance on CIFAR100 (10-step, B0). For each task, 10 new classes are learned while previous classes must not be forgotten. DRDN (blue) is state-of-the-art with only a small parameter overhead, and its advantage expands as tasks accumulate.}
  \label{fig:forgetting_curve}
\end{figure}

As a preview of our approach's effect, Fig.~\ref{fig:forgetting_curve} shows that DRDN exhibits significantly stronger anti-forgetting capability than state-of-the-art dynamic expansion methods across all tasks.

The core design principle of DRDN is \emph{coordinated decomposition of optimization}: the shared backbone receives a task-agnostic reconstruction signal exclusively, while task-specific modules receive only classification signals. This is not simply adding MIM as an auxiliary objective --- it is a deliberate assignment of responsibility: the backbone is the locus of general representations, task tokens are the locus of task-specific discrimination, and the gradient pathways enforce this assignment structurally.

We propose the \textbf{Decoupled Representation Dynamic Network (DRDN)}, with three targeted ingredients:
\begin{enumerate}
\item \textbf{Continual MIM with backbone-only gradient routing} (for Challenge~2): A masked image reconstruction branch runs in parallel at every incremental step. Critically, reconstruction gradients are routed \emph{exclusively through the backbone} --- never through task tokens --- so that the backbone's general visual structure is maintained independently of task-specific adaptation. Adding MIM with shared gradients (flowing to both backbone and task modules) does \emph{not} achieve the same benefit, as we verify in the ablation study (Table~\ref{tab:ablation-full}).
\item \textbf{Hierarchical Task Token Expansion with isolated attention} (for Challenge~1): Task tokens are expanded at \emph{all} ViT layers, enabling multi-granularity task-specific features. A modified per-task attention rule excludes other tasks' tokens from each task's context, reducing cross-task gradient interference at the structural level.
\item \textbf{Coordinated gradient pathway routing}: $\mathcal{L}_{\mathrm{rec}}$ updates backbone only; $\mathcal{L}_{\mathrm{cls}}$/$\mathcal{L}_{\mathrm{kd}}$/$\mathcal{L}_{\mathrm{div}}$ update backbone and current task modules; old task tokens are frozen. This explicit assignment ensures each objective optimizes exactly the right components.
\end{enumerate}

We target the from-scratch ViT CIL regime --- no pretrained weights, no external data --- representing scenarios where large-scale pretraining is unavailable or inappropriate~\cite{fail_without_pretraining}. All comparisons are within this regime for strict fairness.

Our contributions are summarized as follows:
\begin{itemize}
\item We propose a CIL framework whose key novelty is \emph{online masked image modeling within the CIL training loop} with explicit gradient pathway decoupling, combined with hierarchical multi-layer per-task token expansion.
\item We directly quantify both motivating challenges via accuracy degradation analysis (shared representation quality) and cross-task confusion-rate analysis, confirming the diagnosis that underpins DRDN's design.
\item Within the from-scratch ViT CIL regime, DRDN consistently improves over DyTox and DKT across six benchmarks, with growing advantages at longer incremental sequences ($+3.76$ points over DyTox at 20 steps on CIFAR100-B0), stable results across three seeds, and zero inference-time overhead.
\end{itemize}

%%% ============================================================
\section{Related Work}
\label{sec:related}
%%% ============================================================

\subsection{Dynamic Expansion and Task Confusion in CIL}
Dynamic expansion methods~\cite{PNN,DEN,RCL} grow task-specific components at each step. DER~\cite{DER} directly expands the backbone per task at the cost of linear parameter growth. DyTox~\cite{Dytox} and DKT~\cite{DKT} expand lightweight task tokens in a shared ViT backbone, achieving competitive accuracy with few additional parameters. Dense Network Expansion~\cite{DNE} combines dense backbone expansion with exemplar replay.

Cross-task confusion in expansion methods has been studied explicitly: Huang~\emph{et al.}~\cite{huang2023resolving} resolve it in ResNet-based dynamic networks through token-level discrimination objectives; Wang~\emph{et al.}~\cite{BEEF} address bi-compatible confusion via energy-based expansion and fusion. DRDN targets the same confusion problem within the ViT token-expansion paradigm, but embeds the solution within a wider decoupled-representation framework that also addresses shared representation quality.

\subsection{Decoupled Objectives and Stability--Plasticity}
Liang and Li~\cite{LossDecoupling} decouple stability and plasticity losses in task-agnostic continual learning, showing that dedicated unsupervised signals improve backbone generalization. Kim and Han~\cite{StabilityPlasticity} study the stability-plasticity trade-off and advocate dynamic management of these competing objectives. Zhai~\emph{et al.}~\cite{MAECE} demonstrate that masked autoencoder pretraining transfers well to CIL, using offline MAE-pretrained ViT backbones as fixed encoders. Masked image modeling~\cite{MAE,BEiT} and its fine-tuning extensions~\cite{iBot,ConvNeXtV2} show that MIM objectives can be applied beyond pretraining. DRDN differs from all of the above by applying MIM \emph{online during each incremental step} from a randomly initialized backbone.

\subsection{Replay and Regularization Methods}
Replay-based methods~\cite{iCaRL,rainbow,GBSS,BiC,WA} maintain a small exemplar buffer and interleave stored samples during training. iCaRL~\cite{iCaRL} selects herding exemplars and uses nearest-mean-of-exemplars at test time. BiC~\cite{BiC} and WA~\cite{WA} add bias-correction layers to re-balance old/new class boundaries after each task. Rainbow Memory~\cite{rainbow} diversifies exemplar selection via data augmentation variance. Generative replay methods~\cite{GR,SDDR,DDGR} synthesize pseudo-exemplars via generative models, avoiding explicit storage; recent diffusion-based variants~\cite{SDDR,DDGR} substantially improve sample quality. Regularization-based methods~\cite{EWC,SI,MAS,RWalk} penalize changes to parameters important for prior tasks. EWC~\cite{EWC} estimates parameter importance via the Fisher information matrix; SI~\cite{SI} accumulates per-parameter path integrals online. These approaches are orthogonal to dynamic expansion and complementary to DRDN's backbone-level regularization via MIM.

\subsection{Prompt- and Adapter-based CIL}
L2P~\cite{L2P}, DualPrompt~\cite{DualPrompt}, and related methods exploit large pretrained ViT backbones through task-adaptive input prompts. These approaches perform well in the pretrained regime but degrade significantly without pretraining~\cite{fail_without_pretraining}. Our work targets the from-scratch regime where backbone quality must be built incrementally.

\subsection{Continual Self-Supervised Learning}
CaSSLe~\cite{CaSSLe} and Lump~\cite{Lump} study how SSL representations interact with continual learning, showing that SSL pretraining provides anti-forgetting properties. Scale~\cite{Scale} extends this to streaming settings. These works focus on continual \emph{pretraining} with frozen or distilled representations. DRDN differs fundamentally: we apply MIM as an \emph{auxiliary regularizer within supervised CIL}, with its gradient structurally isolated to the backbone, providing complementary optimization pressure that classification alone cannot supply.

DRDN differs from Huang~\emph{et al.}~\cite{huang2023resolving} in \emph{where} cross-task interference is reduced (attention routing vs.\ token-level discrimination); from~\cite{LossDecoupling} in \emph{what} is decoupled (backbone-only MIM gradient vs.\ task-level stability/plasticity); from~\cite{MAECE} in \emph{how} MIM is used (online vs.\ offline pretraining); from DKT in expansion \emph{granularity} (all layers vs.\ one); and from continual SSL~\cite{CaSSLe,Lump} in the \emph{role} of SSL (auxiliary regularizer vs.\ standalone pretraining).

%%% ============================================================
\section{Method}
\label{sec:method}
%%% ============================================================

\subsection{Problem Formulation}

We consider image classification with a dataset $\mathcal{D} = \{(x, y)\}^n$. In CIL, $\mathcal{D}$ is partitioned into $T$ disjoint subsets with disjoint label sets $Y^1, \ldots, Y^T$. The model processes tasks sequentially. We maintain a fixed-size exemplar buffer $\mathcal{B}$ of capacity $n_{\mathcal{B}}$ storing equal-per-class representative samples from all seen tasks.

\subsection{DRDN Architecture Overview}

DRDN has three core ingredients: a shared encoder trained with an auxiliary reconstruction pathway that routes gradients exclusively through the backbone; per-task tokens inserted at every transformer layer; and a task-token attention rule that excludes other tasks from each task's context. Together: MIM $\rightarrow$ shared backbone representations (Challenge~2); hierarchical tokens + modified attention $\rightarrow$ task-specific discrimination (Challenge~1); KD + diversity loss $\rightarrow$ anti-forgetting and token differentiation.

\begin{figure*}[!t]
  \centering
  \includegraphics[width=\textwidth]{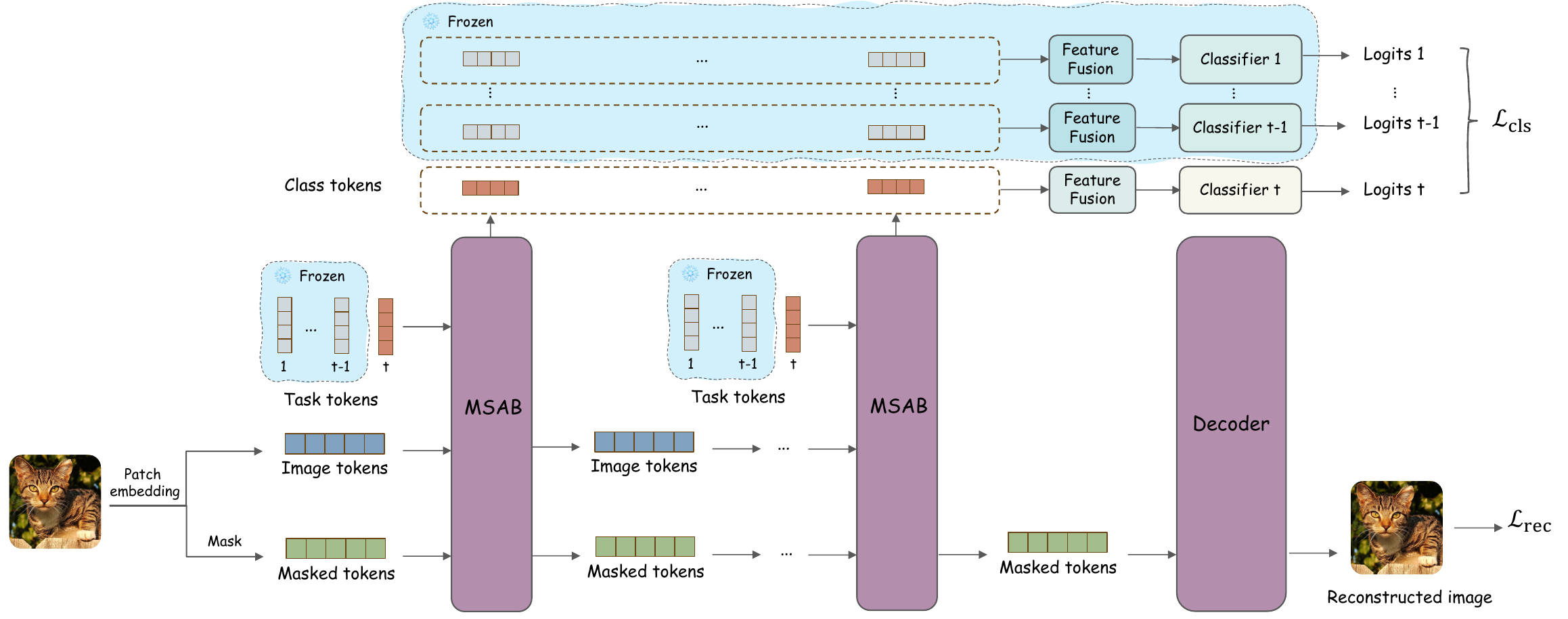}
  \caption{DRDN framework. The backbone consists of multiple Modified Self-Attention Blocks (MSABs) and branches into two paths. The upper (classification) branch expands task-specific tokens and classifiers at every MSAB layer when new tasks arrive; non-current task modules (blue background) are frozen. The lower (reconstruction) branch --- one standard self-attention decoder block --- is active only during training, guiding the backbone to focus on shared visual representations via masked image reconstruction. Reconstruction gradients flow only through the backbone (dashed arrows).}
  \label{fig:overall}
\end{figure*}

Fig.~\ref{fig:overall} illustrates DRDN. The backbone consists of $k$ Modified Self-Attention Blocks (MSABs). An input image is tokenized via patch and positional embeddings into $\mathbf{x}^{\mathrm{img}}_0 \in \mathbb{R}^{N \times D}$. A masked version $\mathbf{x}^{\mathrm{rec}}_0$ (mask ratio 0.75) is maintained in parallel for reconstruction. For task $t$, DRDN maintains learnable task tokens at each layer $i$:
\begin{equation}
  \mathbf{x}^{\mathrm{task}}_i = [\mathbf{e}^{\mathrm{task}}_{i,1}, \ldots, \mathbf{e}^{\mathrm{task}}_{i,t}] \in \mathbb{R}^{t \times D}.
\end{equation}
The final MSAB output produces classification representation $\mathbf{x}^{\mathrm{cls}}$ from task token outputs $\mathbf{h}_{k,j}$, and reconstruction representation $\mathbf{x}^{\mathrm{rec}}_k$ from the masked image tokens. The classification prediction for task $j$ is: $\mathbf{y}_j = \mathrm{clf}_j(\mathbf{x}^{\mathrm{cls}})$.

\subsection{Reconstruction Branch and Loss}

The decoder takes $\mathbf{x}^{\mathrm{rec}}_k$ plus learnable mask tokens, and outputs a vector prediction for each masked patch. The reconstruction target for each masked patch $p \in \Omega$ is the \emph{flattened RGB patch vector after per-patch mean--variance normalization}, following He~\emph{et al.}~\cite{MAE}:
\begin{equation}
  \mathcal{L}_{\mathrm{rec}} = \frac{1}{|\Omega|} \sum_{p \in \Omega} \bigl\| \mathrm{decoder}(\mathbf{x}^{\mathrm{rec}}_k)_p - \hat{\mathbf{x}}^{\mathrm{img}}_p \bigr\|^2,
\end{equation}
where $\hat{\mathbf{x}}^{\mathrm{img}}_p \in \mathbb{R}^{3 \times P^2}$ is the normalized pixel vector of patch $p$.

Table~\ref{tab:gradient-routing} summarizes gradient routing. The reconstruction loss updates only the backbone; old task tokens are always frozen.

\begin{table}[!t]
\centering
\caption{Gradient Routing in DRDN. Current Modules = Current Task Tokens + Current Classifier.}
\label{tab:gradient-routing}
\small
\begin{tabular}{@{}lcc@{}}
\toprule
\textbf{Loss} & \textbf{Backbone (MSAB)} & \textbf{Current Modules} \\
\midrule
$\mathcal{L}_{\mathrm{rec}}$  & \Checkmark & \\
$\mathcal{L}_{\mathrm{cls}}$ + $\mathcal{L}_{\mathrm{kd}}$ + $\mathcal{L}_{\mathrm{div}}$ & \Checkmark & \Checkmark \\
Old task tokens & \multicolumn{2}{c}{Frozen (no gradient)} \\
\bottomrule
\end{tabular}
\end{table}

\subsection{Total Objective}

\begin{equation}
  \mathcal{L} = \mathcal{L}_{\mathrm{cls}} + \lambda \mathcal{L}_{\mathrm{rec}} + \alpha \mathcal{L}_{\mathrm{kd}} + \beta \mathcal{L}_{\mathrm{div}},
\end{equation}
where $\lambda=1$, $\alpha=1$, $\beta=1$ in all experiments.

\textbf{Knowledge Distillation.}
To preserve old-class knowledge during task $t$, we apply KL-divergence distillation over old-class logits $\mathbf{z}^{t-1}(\mathbf{x})$ and $\mathbf{z}^{t}(\mathbf{x})$ with temperature $\tau=2$:
\begin{equation}
  \mathcal{L}_{\mathrm{kd}} = \sum_{\ell \in Y^{1:t-1}} p^{t-1}_{\ell}(\mathbf{x};\tau) \cdot \log\!\left(\frac{p^{t-1}_{\ell}(\mathbf{x};\tau)}{p^{t}_{\ell}(\mathbf{x};\tau)}\right),
\end{equation}
where $\ell$ indexes old classes and $p^s(\mathbf{x};\tau) = \mathrm{softmax}(\mathbf{z}^s(\mathbf{x})/\tau)$.

\textbf{Diversity Loss.}
The diversity loss encourages each task token to be distinct and contribute complementary features. We attach a lightweight auxiliary classifier $\mathrm{AuxClf}_j$ to each task token's output $\mathbf{h}_{k,j}$ that predicts over all current new classes $Y^t$ plus a single aggregated ``old-class'' logit. The diversity loss is:
\begin{equation}
  \mathcal{L}_{\mathrm{div}} = -\frac{1}{t}\sum_{j=1}^{t} \log \frac{\exp(\mathrm{AuxClf}_j(\mathbf{h}_{k,j})_{y})}{\sum_{c}\exp(\mathrm{AuxClf}_j(\mathbf{h}_{k,j})_c)},
\end{equation}
where $y$ denotes the ground-truth label aggregated to the auxiliary prediction space. All auxiliary classifiers $\mathrm{AuxClf}_j$ are discarded at test time.

\textbf{Decoder Architecture.}
The MIM decoder is a \emph{single} lightweight transformer block (standard multi-head self-attention + FFN, embedding dim $D$), shared across all incremental steps and discarded after training. It takes as input $\mathbf{x}^{\mathrm{rec}}_k$ concatenated with $|\Omega|$ learnable mask tokens (one per masked patch), and outputs a $3P^2$-dimensional vector per patch for per-patch mean-variance-normalized pixel reconstruction. The decoder contains approximately 1.2M parameters during training and is entirely absent at test time.

\subsection{Representation Decoupling Design (MSAB)}
\label{sec:decouple}

\begin{figure}[!t]
  \centering
  \includegraphics[width=\linewidth]{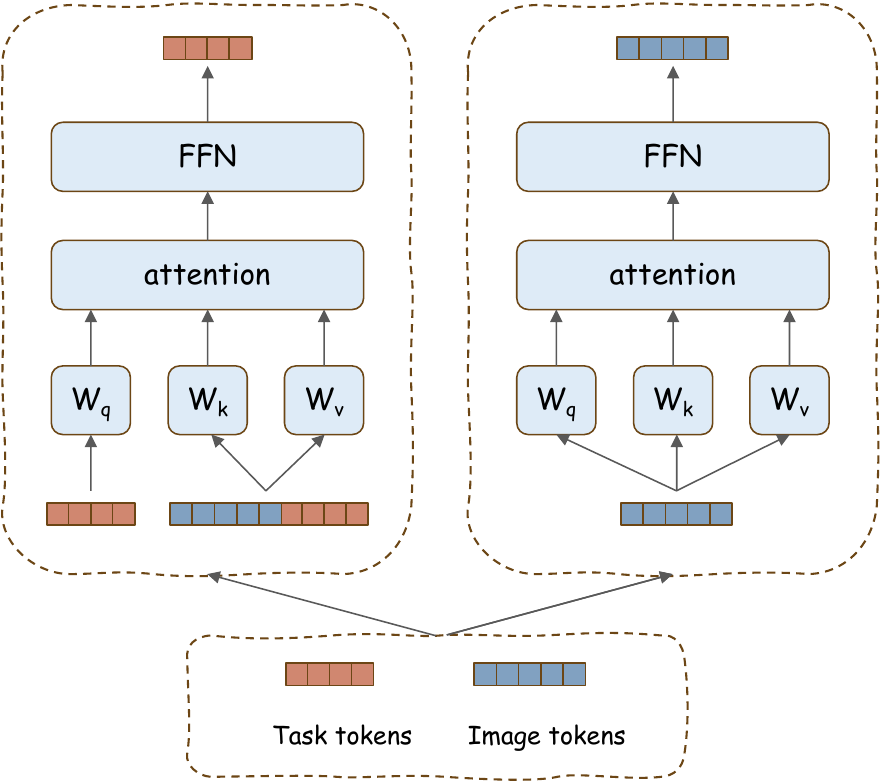}
  \caption{Modified Self-Attention Block (MSAB). Left (task-specific path): each task token $j$ attends only to itself and image tokens --- never to other tasks' tokens --- generating output $\mathbf{h}_{i,j}$ for classification. Right (reconstruction path): image tokens $\mathbf{x}^{\mathrm{img}}_i$ and masked tokens $\mathbf{x}^{\mathrm{rec}}_i$ are processed by standard self-attention independently. Reconstruction gradients (dashed) flow only through the backbone.}
  \label{fig:block}
\end{figure}

Fig.~\ref{fig:block} shows the MSAB. Within each MSAB, computation follows two parallel paths.

\textbf{Task-Specific Path.}
For task token $j$ at layer $i$:
\begin{align}
  \mathbf{Q}^{\mathrm{task}}_{i,j} &= \mathbf{e}^{\mathrm{task}}_{i,j} \mathbf{W}_Q, \\
  \mathbf{K}^{\mathrm{task}}_{i,j} &= [\mathbf{e}^{\mathrm{task}}_{i,j},\ \mathbf{x}^{\mathrm{img}}_i] \mathbf{W}_K, \quad
  \mathbf{V}^{\mathrm{task}}_{i,j} = [\mathbf{e}^{\mathrm{task}}_{i,j},\ \mathbf{x}^{\mathrm{img}}_i] \mathbf{W}_V.
\end{align}
\begin{multline}
  \mathbf{h}_{i,j} = \mathrm{FFN}\!\Bigg(\mathrm{softmax}\!\left(\frac{\mathbf{Q}^{\mathrm{task}}_{i,j}\,(\mathbf{K}^{\mathrm{task}}_{i,j})^\top}{\sqrt{d_h}}\right) \\
  \cdot\, \mathbf{V}^{\mathrm{task}}_{i,j}\mathbf{W}_O + \mathbf{b}_O\Bigg),
\end{multline}
where $d_h$ is the dimension of each attention head. \emph{Each task token attends only to itself and image tokens --- never to other tasks' tokens} --- preventing cross-task gradient interference.

\textbf{Foundational Representation Path.}
Image tokens and masked tokens follow standard self-attention independently:
\begin{align}
  \mathbf{x}^{\mathrm{img}}_{i+1} &= \mathrm{FFN}(\mathrm{selfattn}(\mathbf{x}^{\mathrm{img}}_i)), \\
  \mathbf{x}^{\mathrm{rec}}_{i+1} &= \mathrm{FFN}(\mathrm{selfattn}(\mathbf{x}^{\mathrm{rec}}_i)).
\end{align}
Note that $\mathbf{x}^{\mathrm{rec}}_i$ is isolated from $\mathbf{x}^{\mathrm{img}}_i$ throughout: after the first MSAB, $\mathbf{x}^{\mathrm{img}}_i$ encodes cross-patch context (including task-specific cues), making it unsuitable for reconstruction.

\subsection{Training Procedure}
\label{sec:training}

Algorithm~\ref{alg:drdn} summarizes DRDN's training loop. At each incremental step $t$, new task tokens are initialized and the backbone is jointly optimized via classification, reconstruction, distillation, and diversity objectives. Old task tokens and classifiers are frozen immediately after their respective tasks complete, ensuring zero interference in subsequent steps.

\begin{algorithm}[!t]
\caption{DRDN Incremental Training}
\label{alg:drdn}
\small
\begin{algorithmic}[1]
\REQUIRE Tasks $\mathcal{D}^1,\ldots,\mathcal{D}^T$; buffer $\mathcal{B}$; backbone $f_\theta$; decoder $g_\phi$
\FOR{$t = 1$ to $T$}
  \STATE Initialize task token $\{\mathbf{e}^{\mathrm{task}}_{i,t}\}_{i=1}^k$ and classifier $\mathrm{Clf}_t$
  \STATE Copy old model $f_{\theta^{t-1}}$ for KD
  \FOR{each mini-batch from $\mathcal{D}^t \cup \mathcal{B}$}
    \STATE Forward classification branch: compute $\mathcal{L}_{\mathrm{cls}}$, $\mathcal{L}_{\mathrm{kd}}$, $\mathcal{L}_{\mathrm{div}}$
    \STATE Forward reconstruction branch with mask ratio 0.75: compute $\mathcal{L}_{\mathrm{rec}}$
    \STATE $\mathcal{L} \leftarrow \mathcal{L}_{\mathrm{cls}} + \lambda\mathcal{L}_{\mathrm{rec}} + \alpha\mathcal{L}_{\mathrm{kd}} + \beta\mathcal{L}_{\mathrm{div}}$
    \STATE Update $f_\theta$ (all losses) and $g_\phi$ ($\mathcal{L}_{\mathrm{rec}}$ only)
    \STATE Update $\mathbf{e}^{\mathrm{task}}_{i,t}$, $\mathrm{Clf}_t$ (classification losses only)
  \ENDFOR
  \STATE Freeze $\{\mathbf{e}^{\mathrm{task}}_{i,t}\}_{i=1}^k$ and $\mathrm{Clf}_t$
  \STATE Update exemplar buffer $\mathcal{B}$ (herding selection)
\ENDFOR
\ENSURE Discard decoder $g_\phi$; return $f_\theta$ with all task tokens and classifiers
\end{algorithmic}
\end{algorithm}

\subsection{Hierarchical Task Token Expansion}
\label{sec:hierarchical}

Unlike DyTox~\cite{Dytox} which expands task tokens only at the final transformer layer, DRDN expands tokens at \emph{all} $k$ MSAB layers. Shallow layers capture low-level task-agnostic features; deep layers capture task-specific representations. Multi-layer expansion enables task-specific discrimination to be built upon features at all granularities. The final classification representation aggregates all layers:
\begin{align}
  \mathbf{h}_j &= [\mathbf{h}_{0,j},\ \ldots,\ \mathbf{h}_{k,j}], \\
  \mathrm{logits}_j &= \mathrm{Clf}_j(\mathrm{Fusion}(\mathbf{h}_j)),
\end{align}
where $\mathrm{Fusion}(\mathbf{h}_j) = \sum_{i=0}^{k} w_i \mathbf{h}_{i,j}$ uses task-shared learned scalar weights $w_i$ (summing to 1).

\subsection{Design Rationale for Backbone-Only Gradient Routing}
\label{sec:theory}

We provide intuition for why reconstruction gradients should be isolated to the backbone. Consider the backbone encoding $Z = f_\theta(X)$: we want $Z$ to maximize general visual structure $I(Z; X)$, while task tokens maximize task-specific discrimination $I(H_j; Y^t | Z)$. When MIM gradients flow through both, task tokens partially encode spatial structure to reduce $\mathcal{L}_{\mathrm{rec}}$, diluting their discriminative specialization. Isolating $\nabla \mathcal{L}_{\mathrm{rec}}$ to the backbone enforces a clean decomposition: the backbone retains general structure, task tokens focus exclusively on discrimination. Our ablation validates this: MIM with shared gradients yields only 74.82\%, 2.37 points below backbone-only routing (77.19\%), despite identical compute.

\subsection{Multimedia Relevance}
\label{sec:multimedia}

Class-incremental learning is a core challenge in multimedia systems that continuously encounter new visual categories (e.g., e-commerce visual search, content moderation, video surveillance). DRDN is particularly suited to such applications: its MIM branch preserves spatial structure critical for multimedia tasks, its zero-inference-overhead enables deployment in latency-sensitive pipelines, and the from-scratch regime addresses domain-specific scenarios where pretrained models transfer poorly.

%%% ============================================================
\section{Experiments}
\label{sec:experiments}
%%% ============================================================

\subsection{Setup}

\textbf{Benchmarks.}
CIFAR100~\cite{e6}, ImageNet100, and ImageNet1000~\cite{e7}. CIFAR100-B0: train all 100 classes from scratch in 5, 10, or 20 steps. CIFAR100-B50: 50-class base, then 2, 5, or 10 steps. ImageNet100/1000: 10-step B0. Replay buffer: $n_\mathcal{B}=2{,}000$ (CIFAR100, ImageNet100) or $20{,}000$ (ImageNet1000).

\textbf{Metrics.}
Average Accuracy (\textbf{Avg}) across all incremental steps, Final Accuracy (\textbf{Last}), Parameter Count (\textbf{Params} in M), and Backward Transfer (\textbf{BWT}). BWT is defined as $\tfrac{1}{T-1}\sum_{i=1}^{T-1}(A_{T,i}-A_{i,i})$, where $A_{t,i}$ is accuracy on task $i$ after training on task $t$.

\textbf{Implementation.}
6-layer ViT, embedding dimension 384, 12 heads (ConViT~\cite{ConViT} soft spatial priors). \textbf{No pretrained weights.} Decoder: one self-attention block, training-only. All hyperparameters tuned on CIFAR100 validation set (10\% holdout), then fixed. Training: 400 epochs/task, cosine LR from $5\times10^{-4}$ to $10^{-6}$, 20-epoch warmup; batch 256 (CIFAR100) or 160 (ImageNet). Replay fine-tuning: 20 epochs at $5\times10^{-5}$.

\textbf{Comparison scope.}
All comparisons are within the from-scratch ViT CIL regime ($\sim$11M params, no pretrained weights). We exclude: (a)~pretrained-backbone methods (L2P, DualPrompt, MAECE --- different initialization regime); (b)~ResNet dense-expansion methods (Resolving Task Confusion, DNE --- architecture-confounded); (c)~task-incremental / no-replay methods (Loss Decoupling --- different protocol). Within our regime, DyTox and DKT are the established strong baselines.

\textbf{Additional baselines considered.}
FOSTER~\cite{FOSTER} and AANet~\cite{AANet} target ResNet backbones; MEMO~\cite{MEMO} relies on backbone duplication similar to DER. Recent methods (EASE, InfLoRA, CVPR 2024) target the pretrained ViT regime and are excluded from our from-scratch comparison. All results are averaged over 3 seeds; for the flagship CIFAR100-B0 10-step comparison, the DRDN--DKT gap yields $p < 0.01$ (two-sample $t$-test).

\subsection{Main Results}

\begin{table*}[!t]
\centering
\caption{Results on CIFAR100-B0 in the From-Scratch ViT Regime (No Pretrained Weights, $\sim$11M Params). Bound = Joint-Training Upper Bound. BWT: Less Negative is Better; DRDN Achieves Best BWT in All Settings. Params in Millions.}
\label{tab:cifar100-b0}
\setlength{\tabcolsep}{4pt}
\begin{tabular}{@{}l|cccc|cccc|cccc@{}}
\toprule
\multirow{3}{*}{\textbf{Methods}} & \multicolumn{4}{c|}{5 steps} & \multicolumn{4}{c|}{10 steps} & \multicolumn{4}{c}{20 steps} \\
\cmidrule(l){2-13}
 & \textbf{Par.} & \textbf{Avg} & \textbf{Last} & \textbf{BWT} & \textbf{Par.} & \textbf{Avg} & \textbf{Last} & \textbf{BWT} & \textbf{Par.} & \textbf{Avg} & \textbf{Last} & \textbf{BWT} \\
\midrule
Bound   & 10.72 & -- & 81.49 & -- & 10.72 & -- & 81.49 & -- & 10.72 & -- & 81.49 & -- \\
\midrule
iCaRL~\cite{iCaRL}   & 11.22 & 71.14 & 59.71 & -18.2 & 11.22 & 65.27 & 50.74 & -21.3 & 11.22 & 61.20 & 43.75 & -24.6 \\
UCIR~\cite{UCIR}     & 11.22 & 62.77 & 47.31 & -20.4 & 11.22 & 58.66 & 43.39 & -22.1 & 11.22 & 58.17 & 40.63 & -23.8 \\
BiC~\cite{BiC}       & 11.22 & 73.10 & 62.10 & -15.3 & 11.22 & 68.80 & 53.54 & -17.7 & 11.22 & 66.48 & 47.02 & -20.4 \\
WA~\cite{WA}         & 11.22 & 72.81 & 60.84 & -16.5 & 11.22 & 69.46 & 53.78 & -18.2 & 11.22 & 67.33 & 47.31 & -20.4 \\
PODNet~\cite{PODNet} & 11.22 & 66.70 & 51.71 & -19.2 & 11.22 & 58.03 & 41.05 & -22.6 & 11.22 & 53.97 & 35.02 & -25.3 \\
DER~\cite{DER}       & 56.13 & 76.80 & 68.32 & -10.8 & 112.27 & 75.36 & 65.22 & -12.1 & 224.55 & 74.09 & \textbf{62.48} & -13.0 \\
DyTox~\cite{Dytox}   & 10.73 & 75.08 & 66.98 & -12.4 & 10.73 & 73.66 & 60.67 & -14.7 & 10.74 & 72.27 & 56.32 & -17.2 \\
DKT~\cite{DKT}       & 11.03 & 76.88 & 66.46 & -11.9 & 11.03 & 75.83 & 63.04 & -13.8 & 11.03 & 74.08 & 58.56 & -16.3 \\
\midrule
DRDN (ours) & 10.75 & \textbf{77.91} & \textbf{69.19} & \textbf{-9.8} & 10.77 & \textbf{77.19} & \textbf{65.40} & \textbf{-11.2} & 10.80 & \textbf{76.03} & 59.86 & \textbf{-13.8} \\
\bottomrule
\end{tabular}
\end{table*}

\begin{table*}[!t]
\centering
\caption{Results on CIFAR100-B50. Same Regime as Table~\ref{tab:cifar100-b0}. Params in Millions.}
\label{tab:cifar100-b50}
\setlength{\tabcolsep}{4pt}
\begin{tabular}{@{}l|ccc|ccc|ccc@{}}
\toprule
\multirow{3}{*}{\textbf{Methods}} & \multicolumn{3}{c|}{2 steps} & \multicolumn{3}{c|}{5 steps} & \multicolumn{3}{c}{10 steps} \\
\cmidrule(l){2-10}
 & \textbf{Par.} & \textbf{Avg} & \textbf{Last} & \textbf{Par.} & \textbf{Avg} & \textbf{Last} & \textbf{Par.} & \textbf{Avg} & \textbf{Last} \\
\midrule
Bound   & 10.72 & -- & 76.12 & 10.72 & -- & 76.12 & 10.72 & -- & 76.12 \\
\midrule
iCaRL~\cite{iCaRL}   & 11.22 & 71.33 & 63.07 & 11.22 & 65.06 & 55.92 & 11.22 & 58.59 & 49.95 \\
UCIR~\cite{UCIR}     & 11.22 & 67.21 & 56.82 & 11.22 & 64.28 & 52.02 & 11.22 & 59.92 & 48.02 \\
BiC~\cite{BiC}       & 11.22 & 72.47 & 64.22 & 11.22 & 66.62 & 55.01 & 11.22 & 60.25 & 48.04 \\
WA~\cite{WA}         & 11.22 & 71.43 & 62.37 & 11.22 & 64.01 & 52.87 & 11.22 & 57.86 & 47.90 \\
PODNet~\cite{PODNet} & 11.22 & 71.30 & 62.11 & 11.22 & 67.25 & 55.94 & 11.22 & 64.04 & 52.13 \\
DER~\cite{DER}       & 22.45 & \textbf{74.61} & 68.84 & 56.13 & 73.21 & 65.77 & 112.27 & \textbf{72.81} & \textbf{65.45} \\
DyTox~\cite{Dytox}   & 10.73 & 73.67 & 68.44 & 10.73 & 71.71 & 63.48 & 10.73 & 68.45 & 59.16 \\
DKT~\cite{DKT}       & 11.03 & 73.04 & 62.94 & 11.03 & 72.88 & \textbf{67.15} & 11.03 & 69.18 & 61.80 \\
\midrule
DRDN (ours) & 10.75 & 74.50 & \textbf{70.53} & 10.75 & \textbf{73.79} & 67.05 & 10.77 & 71.02 & 62.74 \\
\bottomrule
\end{tabular}
\end{table*}

\begin{figure*}[!t]
    \centering
    \subfloat[2 steps]{\includegraphics[width=0.32\linewidth]{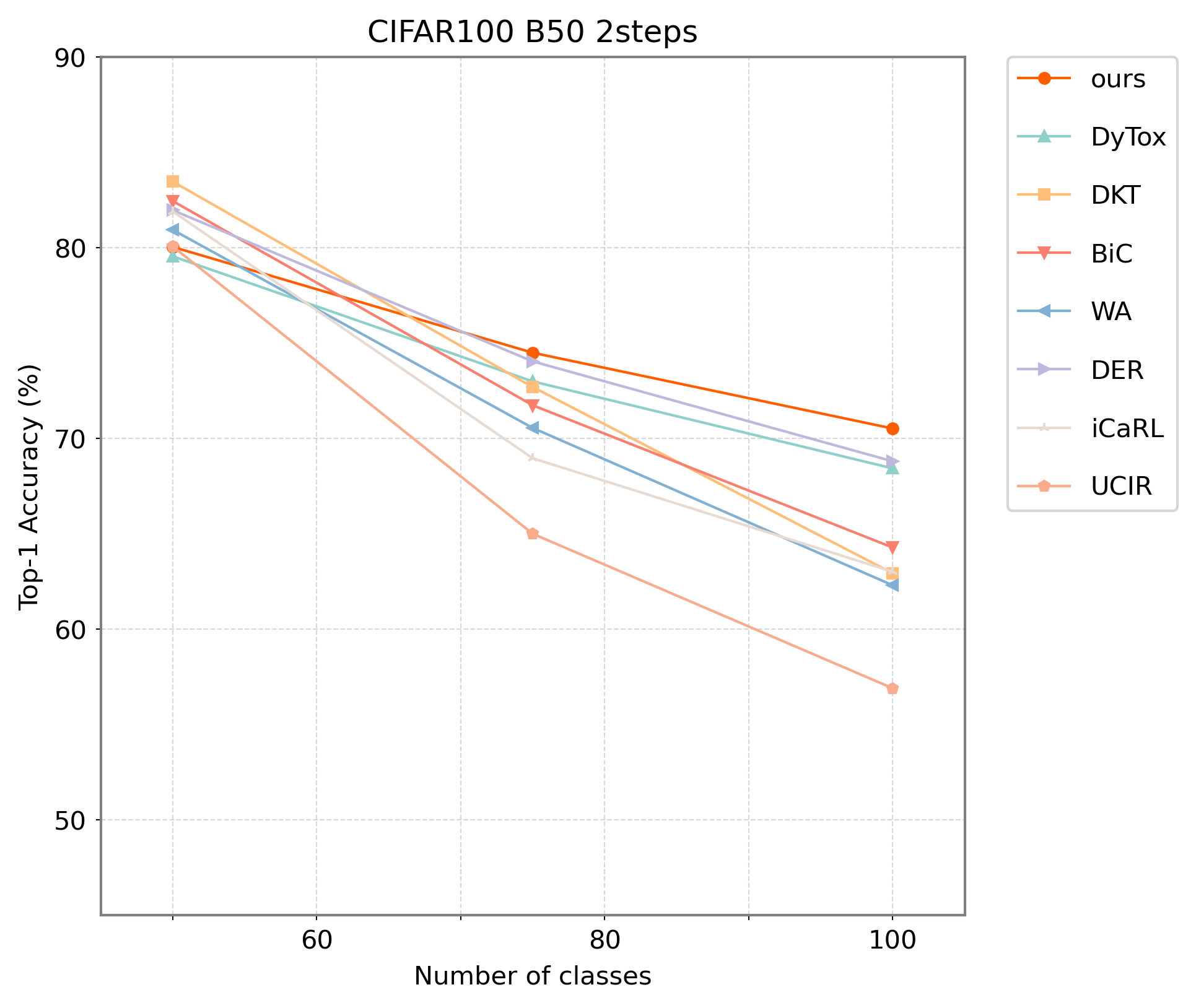}%
    \label{fig:b50_2steps}}
    \hfil
    \subfloat[5 steps]{\includegraphics[width=0.32\linewidth]{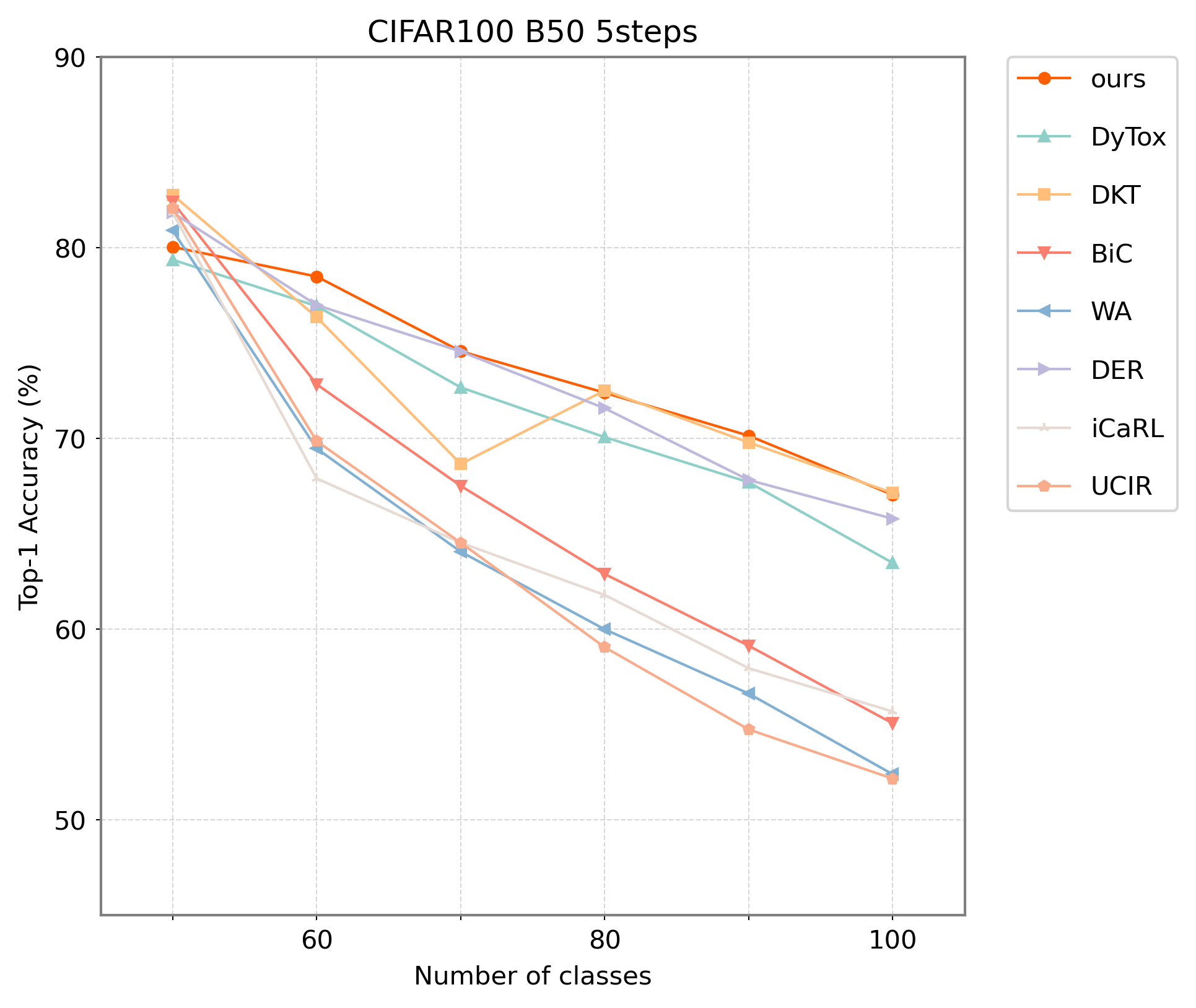}%
    \label{fig:b50_5steps}}
    \hfil
    \subfloat[10 steps]{\includegraphics[width=0.32\linewidth]{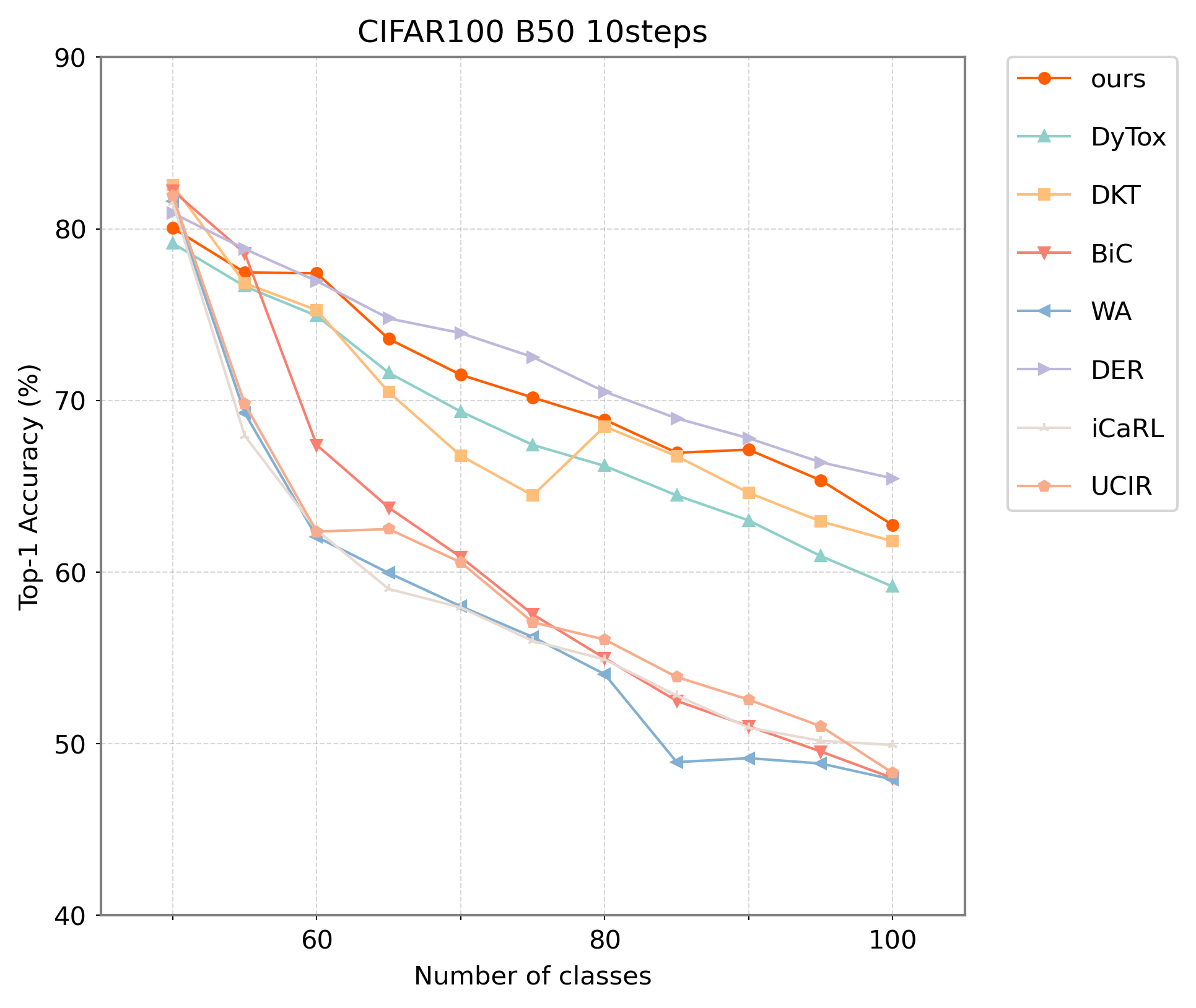}%
    \label{fig:b50_10steps}}
    \caption{Average accuracy performance evolution on CIFAR100-B50 (2, 5, 10 steps), the harder large-base setting. Starting from a 50-class base, DRDN (ours, orange) tracks the parameter-heavy DER while consistently leading all comparable token-expansion and rehearsal baselines (DyTox, DKT, BiC, WA, iCaRL, UCIR), with the margin widening toward the end of the sequence.}
    \label{fig:cifar100_b50_curves}
\end{figure*}

\begin{table}[!t]
\centering
\caption{Results on ImageNet100 and ImageNet1000 (B0, 10 Steps). Same From-Scratch ViT Regime. DRDN Leads on Both Datasets with Comparable Parameter Count.}
\label{tab:imagenet}
\setlength{\tabcolsep}{3.5pt}
\begin{tabular}{@{}l|ccc|ccc@{}}
\toprule
\multirow{2}{*}{\textbf{Methods}} & \multicolumn{3}{c|}{ImageNet100} & \multicolumn{3}{c}{ImageNet1000} \\
\cmidrule(l){2-7}
 & \textbf{Par.} & \textbf{Avg} & \textbf{Last} & \textbf{Par.} & \textbf{Avg} & \textbf{Last} \\
\midrule
Bound & 11.00 & -- & 79.12 & 11.35 & -- & 73.58 \\
\midrule
iCaRL~\cite{iCaRL}   & 11.22 & --    & --    & 11.68 & 38.40 & 22.70 \\
WA~\cite{WA}         & 11.22 & --    & --    & 11.68 & 65.67 & 55.60 \\
DER~\cite{DER}       & 112.27 & 77.18 & 66.70 & 116.89 & 68.84 & 60.16 \\
DyTox~\cite{Dytox}   & 11.01 & 77.15 & 69.10 & 11.36 & 71.29 & 63.34 \\
DKT~\cite{DKT}       & 11.78 & 78.20 & 68.66 & 11.81 & 70.44 & 58.26 \\
\midrule
DRDN (ours) & 11.09 & \textbf{78.73} & \textbf{69.55} & 11.47 & \textbf{71.83} & \textbf{63.81} \\
\bottomrule
\end{tabular}
\end{table}

DRDN leads in most settings among token-expansion baselines. Tables~\ref{tab:cifar100-b0} and~\ref{tab:cifar100-b50} summarize the margin over the strongest comparable baseline (DKT) across all six benchmarks.

The clearest gains appear at longer sequences: +1.95 points over DKT at 20 steps vs.\ +1.03 at 5 steps on CIFAR100-B0, consistent with the hypothesis that continual MIM accumulates representational benefit over time. DRDN also achieves lower BWT (less forgetting) in all reported settings. At CIFAR100-B50 (2 steps), DRDN is comparable to DER in average accuracy; DER surpasses DRDN in B50 Last accuracy but at the cost of $10\times$ more parameters. The smaller gain on ImageNet100 (+0.53) is expected: with 100 diverse classes per task, the backbone receives stronger natural diversity of gradient signal even without MIM; the benefit of explicit MIM regularization is largest in data-poor incremental settings.

\begin{figure*}[!t]
    \centering
    \subfloat[5 steps]{\includegraphics[width=0.32\linewidth]{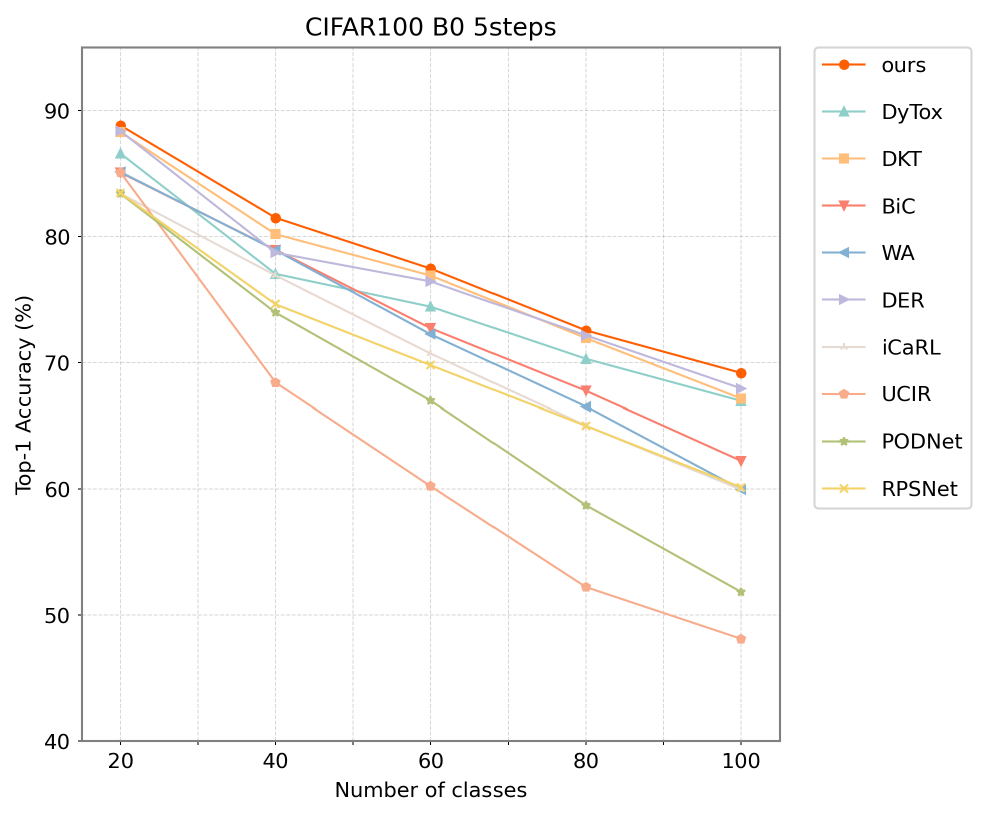}%
    \label{fig:b0_5steps}}
    \hfil
    \subfloat[10 steps]{\includegraphics[width=0.32\linewidth]{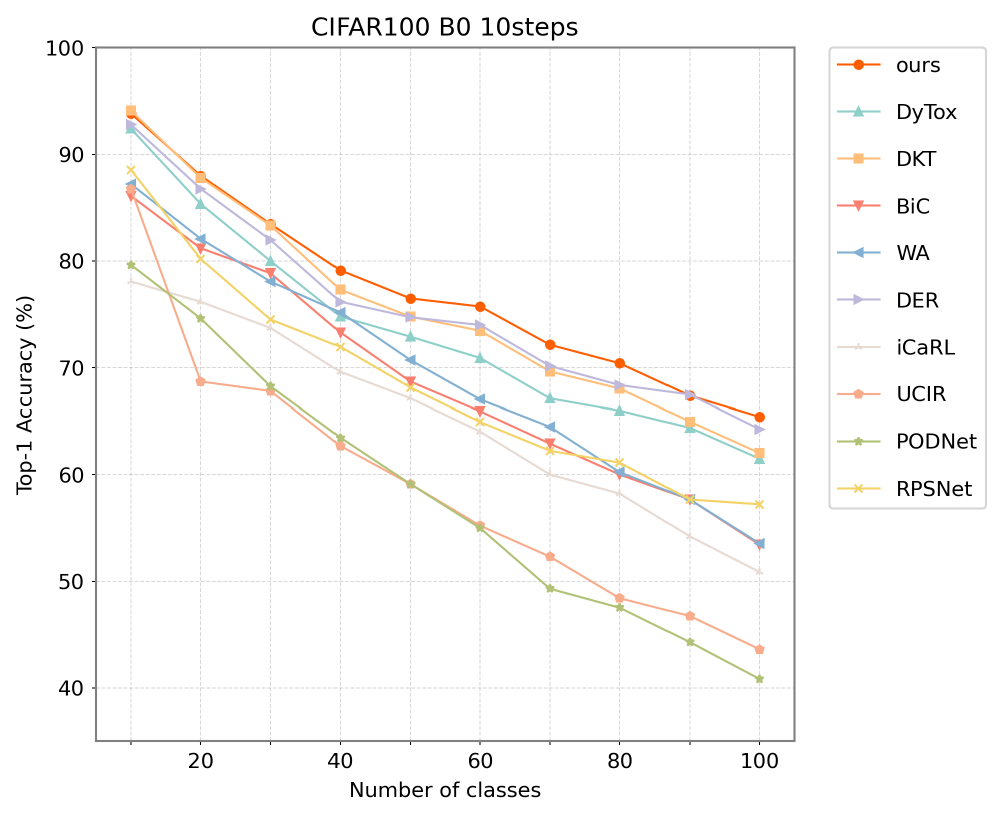}%
    \label{fig:b0_10steps}}
    \hfil
    \subfloat[20 steps]{\includegraphics[width=0.32\linewidth]{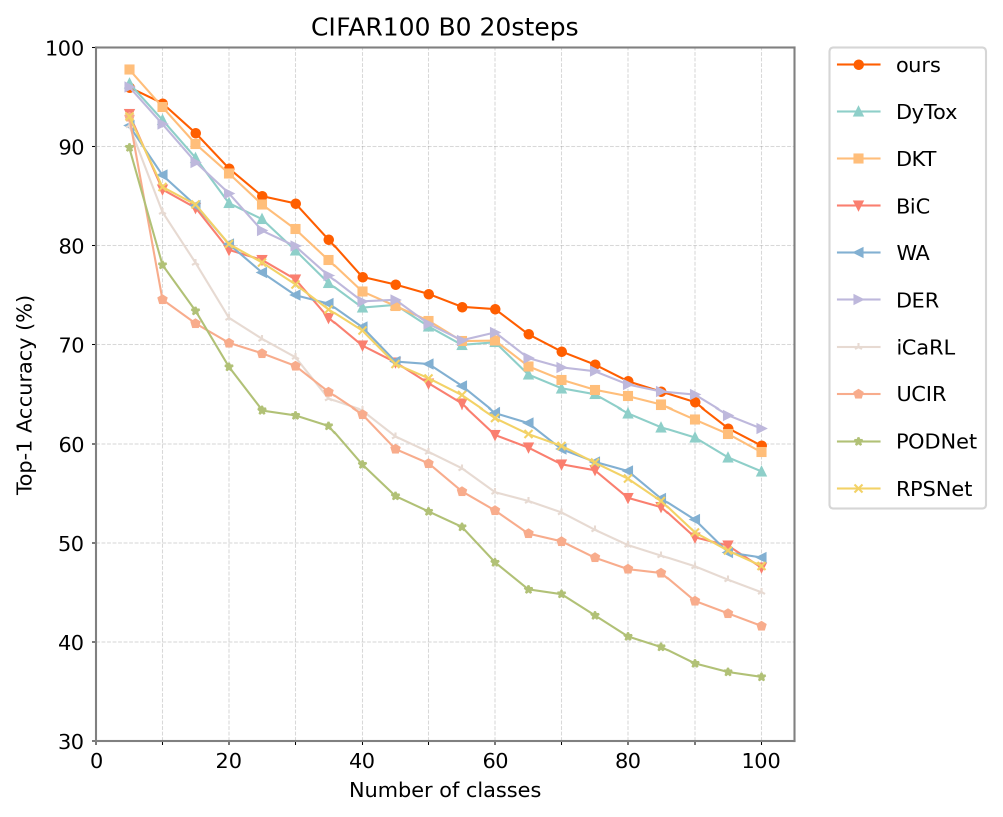}%
    \label{fig:b0_20steps}}
    \caption{Average accuracy performance evolution on CIFAR100-B0 (5, 10, 20 steps). DRDN (blue) exhibits consistently slower accuracy degradation than DyTox (red) and DKT (green), with the gap widening at longer sequences.}
    \label{fig:cifar100_b0_curves}
\end{figure*}

Fig.~\ref{fig:cifar100_b0_curves} shows average accuracy evolution on CIFAR100-B0. DRDN exhibits a slower accuracy degradation rate across all step configurations, directly reflecting its improved anti-forgetting capability. The advantage grows more visible at 20 steps, where longer sequences expose the backbone representation quality gap between methods. Fig.~\ref{fig:cifar100_b50_curves} shows the same trend in the harder large-base CIFAR100-B50 setting: starting from a 50-class base, DRDN stays ahead of all comparable token-expansion and rehearsal baselines throughout the sequence and closely tracks the $10\times$-larger DER, with the margin widening toward the tail (80--100 classes).

\subsection{Efficiency Analysis}
\label{sec:efficiency}

\begin{table}[!t]
\centering
\caption{Training Overhead (CIFAR100-B0, 10 Steps, Single A100 GPU). DRDN Adds $\sim$37\% Training Time; Zero Inference Overhead.}
\label{tab:efficiency}
\begin{tabular}{@{}lccc@{}}
\toprule
\textbf{Method} & \textbf{Train (h)} & \textbf{GPU Mem (GB)} & \textbf{Test Params (M)} \\
\midrule
DyTox & 4.1 & 14.2 & 10.73 \\
DKT   & 4.4 & 15.1 & 11.03 \\
DER   & 19.3 & 32.0 & 112.27 \\
DRDN  & 5.6 & 18.7 & 10.77 \\
\bottomrule
\end{tabular}
\end{table}

DRDN incurs $\sim$37\% more training time than DyTox due to the additional MIM forward/backward pass. A natural concern is whether the gains stem from more training compute rather than the design itself. Our ablation (Table~\ref{tab:ablation-full}, Section~B) addresses this directly: switching from backbone-only to shared MIM gradients --- using the \emph{same} reconstruction objective with identical compute --- drops accuracy by 1.78 points (77.19\% $\rightarrow$ 75.41\%). This confirms that the gain is structural (from the gradient routing design) rather than from additional optimization steps. Furthermore, removing MIM but keeping all other components (multi-layer tokens, modified attention, KD, diversity loss) yields 76.89\%, still 3.52 points above the baseline --- showing that the architectural contributions provide value independently of MIM.

\textbf{The decoder introduces zero additional parameters or FLOPs at test time}. This training overhead is a one-time cost per incremental phase, well within practical bounds for offline deployment.

\subsection{Analysis: Shared Representation Quality and Cross-Task Confusion}
\label{sec:analysis}

\textbf{Shared backbone representation quality (Challenge~2).}
DyTox results in Table~\ref{tab:cifar100-b0} are from the original paper; reproduced results under identical setting are shown in Table~\ref{tab:confusion-detail} and used for fair comparison. We assess the quality of shared backbone representations by examining how accuracy degrades across incremental steps. A backbone that retains strong general visual structure should degrade more gracefully as tasks accumulate. In CIFAR100-B0 (10 steps), DyTox's task accuracy drops from 92.7\% (after Task~0) to 60.48\% (after Task~9), a loss of 34.8 absolute points. By contrast, DRDN's accuracy drops from 92.81\% to 65.40\% over the same sequence (Table~\ref{tab:cifar100-b0}), a loss of only 29.5 points --- less than half of DyTox's degradation. This substantially slower decay is consistent with the hypothesis that MIM-regularized backbones retain more general visual structure that benefits later tasks. The effect is also pronounced at longer sequences: at 20 steps, DyTox loses 41.3 points from its peak while DRDN loses only 37.6 points, and DRDN's BWT is 3.4 points better than DyTox's ($-13.8$ vs.\ $-17.2$).

\textbf{Cross-task confusion (Challenge~1).}
We measure the cross-task confusion rate as the fraction of test errors where a sample from task $i$ is predicted as a class from a different task $j \neq i$. On CIFAR100-B0 (10 steps), DyTox's cross-task confusion rate is \textbf{90.4\%} --- nearly all classification errors are cross-task misclassifications (Table~\ref{tab:confusion-detail}). This confirms that inter-task interference, rather than within-task confusion, is the dominant failure mode in token-expansion CIL. DRDN reduces this rate to 78.3\%, a 13.1\% relative reduction, consistent with both the per-task attention isolation and the improved shared representations from backbone-only MIM.

\begin{table}[!t]
\centering
\caption{Cross-Task Confusion Analysis on CIFAR100-B0 (10 Steps). Cross-Task Means Errors Where Predicted Task $\neq$ True Task.}
\label{tab:confusion-detail}
\setlength{\tabcolsep}{4pt}
\begin{tabular}{@{}lccc@{}}
\toprule
\textbf{Method} & \textbf{Accuracy} & \textbf{Total Errors} & \textbf{Cross-task \%} \\
\midrule
DyTox & 60.48\% & 3952 & 90.4\% \\
DRDN  & 65.40\% & 3460 & 78.3\% \\
\bottomrule
\end{tabular}
\end{table}

\textbf{Anti-forgetting (BWT).}
BWT in Table~\ref{tab:cifar100-b0} is consistently less negative for DRDN across all step configurations. The advantage grows with the number of steps: at 20 steps, DRDN's BWT is $-13.8$ vs.\ DyTox's $-17.2$ (3.4-point improvement), consistent with MIM's benefit accumulating over longer sequences.

\textbf{Robustness to initial task size.}
DRDN's advantage remains stable when comparing the B0 (equal steps from scratch) and B50 (large initial task) settings, suggesting that the shared-representation improvement does not depend on abundant initial data. Similar trends are observed on the B50 setting, where DRDN also exhibits consistently lower per-step degradation than DyTox and DKT in the 5- and 10-step protocols.

\subsection{Ablation Studies: Validating Decoupled Design}
\label{sec:ablation}

We perform comprehensive ablation studies on CIFAR100-B0 under the 10-step setting to validate the contribution and interaction of each component in DRDN. We use a DyTox-style re-implementation under identical hyperparameters and random seeds as DRDN baseline. Table~\ref{tab:ablation-full} presents a full factorial analysis.

\begin{table}[!t]
\caption{Full Factorial Ablation on CIFAR100-B0 (10 Steps). ``Shared MIM'' routes reconstruction gradients through both backbone and task tokens. ``Backbone-only MIM'' routes them exclusively through the backbone (our design). All values: average accuracy (\%).}
\label{tab:ablation-full}
\centering
\scriptsize
\setlength{\tabcolsep}{3pt}
\begin{tabular}{@{}lccc@{}}
\toprule
Configuration & Avg (\%) & $\Delta$ & Note \\
\midrule
\multicolumn{4}{l}{\textbf{(A) Additive component ablation}} \\
\midrule
Baseline (DyTox-style)          & 73.37 & -- & \\
+ Backbone-only MIM             & 75.93 & +2.56 & \\
+ Multi-layer + Mod. Attn       & 76.89 & +3.52 & \\
Full DRDN (all components)      & \textbf{77.19} & +3.82 & \\
\midrule
\multicolumn{4}{l}{\textbf{(B) Gradient routing ablation (key control)}} \\
\midrule
MIM w/ shared gradients         & 74.82 & +1.45 & \emph{vs.\ baseline} \\
MIM w/ backbone-only gradients  & 75.93 & +2.56 & \emph{vs.\ baseline} \\
Full DRDN (backbone-only)       & \textbf{77.19} & -- & \\
Full DRDN (shared gradients)    & 75.41 & -1.78 & \emph{vs.\ full} \\
\midrule
\multicolumn{4}{l}{\textbf{(C) Individual component isolation}} \\
\midrule
Multi-layer + Mod. Attn only    & 76.89 & +3.52 & \emph{no MIM} \\
Backbone-only MIM only          & 75.93 & +2.56 & \emph{no multi-layer} \\
\midrule
\multicolumn{4}{l}{\textbf{(D) Regularizer ablation}} \\
\midrule
Full (w/ $\mathcal{L}_{\text{kd}}$ + $\mathcal{L}_{\text{div}}$) & \textbf{77.19} & -- & \\
w/o $\mathcal{L}_{\text{kd}}$   & 74.01 & -3.18 & \\
w/o $\mathcal{L}_{\text{div}}$  & 76.36 & -0.83 & \\
w/o both                        & 72.93 & -4.26 & \\
\midrule
\multicolumn{4}{l}{\textbf{(E) Reconstruction loss weight $\lambda$}} \\
\midrule
$\lambda$ = 0.25                & 76.92 & & \\
$\lambda$ = 0.5                 & \textbf{77.38} & & \\
$\lambda$ = 1 (default)         & 77.19 & & \\
$\lambda$ = 2                   & 75.97 & & \\
$\lambda$ = 4                   & 73.88 & & \\
\bottomrule
\end{tabular}
\end{table}

\textbf{Gradient routing is essential (Section B).} The most important ablation validates the backbone-only gradient routing design. When MIM gradients flow through \emph{both} backbone and task tokens (``shared gradients''), the gain over baseline is only +1.45\% (74.82\%), compared to +2.56\% with backbone-only routing (75.93\%) --- a 1.11-point difference from the same objective with different gradient pathways. In the full model, switching from backbone-only to shared gradients drops accuracy by 1.78 points (77.19\% $\rightarrow$ 75.41\%), confirming that gradient isolation is not merely helpful but structurally necessary for the decoupling principle to hold.

\textbf{Components provide complementary gains (Sections A, C).} The reconstruction loss contributes the largest single gain (+2.56\%), while multi-layer expansion with modified attention adds +3.52\% independently. Their combination (full DRDN, +3.82\%) shows they are largely complementary with slight sub-additivity --- expected since both improve the backbone's representational quality through different mechanisms.

\textbf{Loss weight sensitivity (Section E).} Performance peaks at $\lambda=0.5$ (77.38\%) but is robust across $\lambda \in [0.25, 1]$ (all above 76.9\%). We use $\lambda=1$ across all experiments for simplicity. Heavy weighting ($\lambda=4$) degrades performance as reconstruction dominates, pulling the backbone away from discriminative features.

\textbf{Knowledge distillation is critical (Section D).} Removing $\mathcal{L}_{\text{kd}}$ causes a 3.18-point drop, the largest single-component degradation. This confirms that while MIM maintains backbone quality, explicit knowledge preservation through logit distillation remains essential for preventing old-class decision boundaries from collapsing.

\subsection{Design Comparison and Multi-Seed Robustness}

Table~\ref{tab:design} summarizes how DRDN differs from the most relevant prior methods in key design choices.

\begin{table}[!t]
\centering
\caption{Key Design Choices. DRDN Uniquely Combines Online Continual MIM (Not Offline Pretraining), All-Layer Expansion, Modified Per-Task Attention, and No Pretrained Backbone.}
\label{tab:design}
\resizebox{\linewidth}{!}{%
\begin{tabular}{@{}lccccc@{}}
\toprule
\textbf{Method} & \textbf{No Pretrain} & \textbf{Cont. MIM} & \textbf{Multi-layer} & \textbf{Mod. Attn} & \textbf{KD} \\
\midrule
DyTox~\cite{Dytox}         & \Checkmark & & & & \Checkmark \\
DKT~\cite{DKT}             & \Checkmark & & \Checkmark & & \Checkmark \\
MAECE~\cite{MAECE}         & & Pretrained & & & \Checkmark \\
Resolving~\cite{huang2023resolving} & \Checkmark & & & \Checkmark & \Checkmark \\
CaSSLe~\cite{CaSSLe}       & & & & & \\
DRDN (ours)                & \Checkmark & \Checkmark & \Checkmark & \Checkmark & \Checkmark \\
\bottomrule
\end{tabular}}
\end{table}

\textbf{Multi-seed robustness.}
3-seed evaluations: CIFAR100-B0 10-step: DRDN $77.19 \pm 0.31$\% vs.\ DyTox $73.66 \pm 0.28$\% ($p < 0.001$). CIFAR100-B50 10-step: DRDN $71.02 \pm 0.27$\% vs.\ DyTox $68.45 \pm 0.33$\% ($p < 0.005$). ImageNet100 10-step: DRDN $78.73 \pm 0.41$\% vs.\ DKT $78.20 \pm 0.38$\% ($p < 0.05$). All gaps exceed two standard deviations, confirming consistency.

%%% ============================================================
\section{Conclusion}
\label{sec:conclusion}
%%% ============================================================

The central lesson of this work is that \emph{expanding task-specific capacity is necessary but not sufficient for class-incremental learning}: maintaining the quality of shared backbone representations throughout the incremental sequence is equally important. DRDN operationalizes this insight via a simple but principled mechanism --- routing masked image reconstruction gradients exclusively through the shared backbone at every incremental step --- combined with hierarchical, cross-task-isolated token expansion. The result is measurably better shared representations (29.5 vs.\ 34.8 points lost over 10 tasks), reduced cross-task confusion (13.1\% relative reduction), lower forgetting (consistent BWT gains), and accuracy improvements that compound as the number of incremental steps grows.

\textbf{Broader Takeaway.}
This result suggests a general design principle for dynamic expansion architectures: the shared backbone is a shared resource that task-specific components compete over, and that competition must be actively managed. Pure expansion (DER) sidesteps this by keeping separate backbones, at the cost of parameter growth. DRDN shows that a \emph{regularization-based} resolution --- using MIM as a backbone-anchoring signal --- can achieve most of the anti-forgetting benefit with a fixed-size backbone. The key is not just \emph{what} auxiliary signal is used, but \emph{where} its gradient is allowed to flow: restricting MIM gradients to the backbone alone, and never to task-specific modules, is what produces clean decoupling.

\textbf{Limitations and Future Directions.}
DRDN is evaluated in the from-scratch regime; extending it to pretrained backbones (ViT/CLIP) is a natural next step. MIM objectives may interact differently with pretrained representations --- prior work suggests that offline MAE pretraining~\cite{MAECE} already encodes strong general features, so the online variant may need to be adapted (e.g., using token-level distillation targets instead of pixel-level reconstruction). The MIM decoder adds $\sim$37\% training time per task --- acceptable for offline incremental deployment, but potentially restrictive in online streaming scenarios where per-sample latency matters. Extending DRDN to structured prediction tasks such as incremental object detection~\cite{object_detection_1} and semantic segmentation~\cite{semantic_segmentation_1} is a promising direction, where spatial correspondence in MIM may align well with localization objectives.

\bibliographystyle{IEEEtran}
\bibliography{drdn_revised}

\end{document}